\documentclass[runningheads]{llncs}
\usepackage{graphicx}
\usepackage{bm}
\usepackage{amssymb}
\usepackage{cite}
\usepackage{amsmath}
\usepackage{fancyhdr}
\usepackage{multicol}
\usepackage{url}
\usepackage{enumerate}
\usepackage{silence}
\usepackage[hang,small,bf]{caption}
\usepackage[labelformat=empty]{subcaption}
\usepackage{booktabs}
\usepackage{multirow}
\usepackage{color}
\fancypagestyle{normal}
{
\fancyhead{}
\fancyhead[R]{Appendix}
\fancyfoot{}

\cfoot{\thepage}
}

\newcommand{\eref}[1]{Equation~(\ref{#1})}
\newcommand{\fref}[1]{{Fig.~\ref{#1}}}
\newcommand{\tref}[1]{{Table~\ref{#1}}}
\newcommand{\sref}[1]{Section~\ref{#1}}

\newcommand{\Gcenter}[2]{
  \dimen0=\ht\strutbox%
  \advance\dimen0\dp\strutbox%
  \multiply\dimen0 by#1%
  \divide\dimen0 by2%
  \advance\dimen0 by-.5\normalbaselineskip
  \raisebox{-\dimen0}[0pt][0pt]{#2}}

\newcommand{\etal}{et al.}

\begin{document}
\title{High-Quality Virtual Single-Viewpoint Surgical Video: Geometric Autocalibration of Multiple Cameras in Surgical Lights}
\author{Yuna Kato\inst{1} \and
    Mariko Isogawa\inst{1} \and
    Shohei Mori\inst{2,1} \and
    Hideo Saito\inst{1} \and
    \\
    Hiroki Kajita\inst{1} \and
    Yoshifumi Takatsume\inst{1}
}

\authorrunning{Y. Kato \etal}
\institute{
    Keio University \and Graz University of Technology
}
\maketitle              %

\begin{abstract}
Occlusion-free video generation is challenging due to surgeons' obstructions in the camera field of view. Prior work has addressed this issue by installing multiple cameras on a surgical light, hoping some cameras will observe the surgical field with less occlusion. However, this special camera setup poses a new imaging challenge since camera configurations can change every time surgeons move the light, and manual image alignment is required. This paper proposes an algorithm to automate this alignment task. The proposed method detects frames where the lighting system moves, realigns them, and selects the camera with the least occlusion. This algorithm results in a stabilized video with less occlusion. Quantitative results show that our method outperforms conventional approaches. A user study involving medical doctors also confirmed the superiority of our method.

\keywords{Surgical Video Synthesis \and Multi-view Camera Calibration \and Event Detection.}
\end{abstract}

\section{Introduction}
\label{sec:intro}
Surgical videos can provide objective records in addition to medical records. Such videos are used in various applications, including education, research, and information sharing~\cite{example_use,example_education}.
In endoscopic surgery and robotic surgery, the surgical field can be easily captured because the system is designed to place a camera close to it to monitor operations directly within the camera field of view. Conversely, in open surgery, surgeons need to observe the surgical field; therefore, the room for additional cameras can be limited and disturbed \cite{open_surgery}. 

To overcome this issue, Kumar and Pal~\cite{Kumar} installed a stationary camera arm to record surgery. However, their camera system had difficulty recording details (i.e., close-up views) since the camera had to be placed far from the surgical field so as not to disturb the surgeons.
Instead, Nair~\etal~\cite{Nair} used a camera mounted on a surgeon's head, which moved frequently and flexibly.

\begin{figure}[tb]
\includegraphics[width=\hsize]{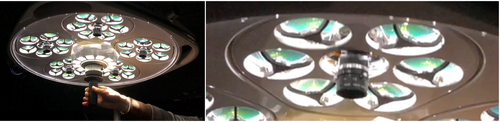}
\caption{Surgical light with multiple cameras. The unit consists of five cameras (left), each of which is surrounded by multiple light sources (right).}
\label{fig:surgicallight}
\vspace{-5mm}
\end{figure}

For solid and stable recordings,
previous studies have installed cameras on surgical lights. 
Byrd~\etal~\cite{Byrd} mounted a camera on a surgical light, which could easily be blocked by the surgeon's head and body. To address this issue, Shimizu~\etal~\cite{Shimizu} developed a surgical light with multiple cameras to ensure that at least one camera would observe the surgical field (\figurename~\ref{fig:surgicallight}). In such a multi-camera system, automatically switching cameras can ensure that the surgical field is visible in the generated video~\cite{Shimizu,hachiuma,saito}.

However, the cameras move every time surgeons move the lighting system, and thus, the image alignment becomes challenging.
Obayashi~\etal~\cite{obayashi} relied on a video player to manually seek and segment a video clip with no camera movement.
This unique camera setup and view-switching approaches create a new task to be fulfilled, which we address in this paper: automatic occlusion-free video generation by automated change detection in camera configuration and multi-camera alignment to smoothly switch to the camera with the least occlusion.\footnote{project page: \color{magenta}https://github.com/isogawalab/SingleViewSurgicalVideo}
In summary, our contributions are as follows: 
\begin{itemize}
\item We are the first to fulfill the task of automatic generation of stable virtual single-view video with reduced occlusion for a multi-camera system installed in a surgical light.
\item We propose an algorithm that detects camera movement timing by measuring the degree of misalignment between the cameras.
\item We propose an algorithm that finds frames with less occluded surgical fields. 
\item We present experiments showing greater effectiveness of our algorithm than conventional methods. 
\end{itemize}

\begin{figure}[tb]
\includegraphics[width=1.0\hsize]{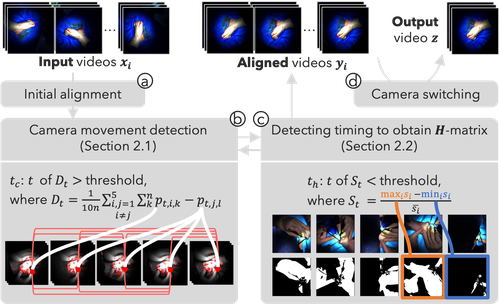}
\vspace{-7mm}
\caption{Overview of the proposed method.}
\label{fig:overview}
\end{figure}
\vspace{-3mm}

\section{Method}
\label{sec:method}
Given a sequence of image frames captured by five cameras installed in a surgical light $\boldsymbol{x}_i=[x_1, x_2, ..., x_T]$ (\figurename~\ref{fig:surgicallight}), our goal is to generate a stable single-view video sequence with less occlusion $\boldsymbol{z}=[z_1, z_2, ..., z_T]$(\fref{fig:overview}). Here, $\boldsymbol{x}_i$ represents a sequence captured with $i$-th camera $c_i$, and $T$ indicates the number of frames.

We perform an initial alignment using the method by Obayashi \etal~\cite{obayashi} (\figurename~\ref{fig:overview}a), which cumulatively collects point correspondences over none-feature rich frames and calculates homography matrices, $M_i$, via a common planar scene proxy.
Then, we iteratively find a frame ID, $t_c$, where the cameras started moving (\sref{sec:21}, \figurename~\ref{fig:overview}b) and a subsequent frame ID, $t_h$, to update homography matrices under no moving cameras and the least occlusion (\sref{sec:22}, \figurename~\ref{fig:overview}c). Updated homography warping can provide a newly aligned camera view sequence $\boldsymbol{y}_i=[y_1, y_2, ..., y_T]$ after the cameras moved.
Finally, using the learning-based object detection method by Shimizu~\etal~\cite{Shimizu}, we select camera views with the least occlusion from $\boldsymbol{y}_i$ (\figurename~\ref{fig:overview}d). Collecting such frames results in a stable single-view video sequence with the least occlusion $\boldsymbol{z}=[z_1, z_2, ..., z_T]$.

\subsection{Camera Movement Detection}\label{sec:21} \par
To find the $t_c$, we use the ``degree of misalignment'' obtained from the sequence $\boldsymbol{y}_i$ of the five aligned cameras. Since the misalignment should be zero if the geometric calibration between the cameras works well and each view overlaps perfectly, it can be used as an indication of camera movement.

First, the proposed method performs feature point detection in each frame of $\boldsymbol{y}_i$. Specifically, we use the SIFT algorithm~\cite{SIFT}. Then, feature point matching is performed for each of the 10 combinations of the five frames. The degree of misalignment $D_t$ at frame $t$ is represented as
\begin{equation}
D_t = \frac{1}{10n}\sum_{\substack{i, j=1 \\ i\neq{j}}}^{5}\sum_{k}^{n}{\bm{p}_{t,i,k} - \bm{p}_{t,j,l}},
\label{equ:score2}
\end{equation}
where $\bm{p}$ and $k$ denote a keypoint position and its index in the $i$-th camera's coordinates respectively, $l$ represents the corresponding index of $k$ in the $j$-th camera's coordinates, and $n$ represents the total number of corresponding points.

If $D_t$ exceeds a certain threshold, our method detects camera movement. However, the calculated misalignment is too noisy to be used as is. To eliminate the noise, the outliers are removed, and smoothing is performed by calculating the movement average. Moreover, to determine the threshold, sample clustering is performed according to the degree of misalignment. Assuming that the multi-camera surgical light never moves more than twice in 10 minutes, the detected degree of misalignment is divided into two classes, one for every 10 minutes. 

The camera movement detection threshold is expressed by \eref{equ:th}, where $t$ represents the frames classified as the frames before the camera movement.
The frame at the moment when the degree of misalignment exceeds the threshold is considered as the frame when the camera moved.
\begin{equation}
threshold=\min{(\max_{t}{(\bm{D}_{t})} + 1,  
        \frac{2}{T}\sum_{t=1}^{T}{\bm{D}_{t}} )}
\label{equ:th}
\end{equation}
To make the estimation of $t_c$ more robust, this process is performed multiple times on the same group of frames, and the median value is used as $t_c$. This is expected to minimize false detections.

\subsection{Detecting the Timing for Obtaining a Homography Matrix}\label{sec:22} \par
$t_h$ represents the timing when to obtain homography matrix. Although it would be ideal to generate always-aligned camera sequences by performing homography transformation on every frame, this would incur high computational costs if the homography is constantly calculated. Therefore, the proposed method calculates the homography matrix only after the cameras have stopped moving.

Unlike a previous work that determined the timing for performing homography transformation manually~\cite{obayashi}, our method automatically detects $t_h$ by using the area of surgical field appearing in surgical videos as an indication. This region indicates the extent of occlusion. Since the five cameras capture the same surgical field, if there is no occluded camera, the area of surgical field will have the same extent in all five camera frames. 
\eref{equ:s} is used to calculate the degree to which the area of surgical field is the same in all five camera frames, where $\bm{s_i}$ is the area of surgical field detected in each camera. 
\begin{equation}
S = \{\max_{i}(\bm{s_i}) - \min_{i}(\bm{s_i})\}/\bm{\bar{s_i}},
\label{equ:s}
\end{equation}
where $S$ is calculated every 30 frames, and if it is continuously below a given threshold (0.5 in this method), the corresponding timing is selected as the $t_h$.

\begin{figure}[tb]
\includegraphics[width=1.0\hsize]{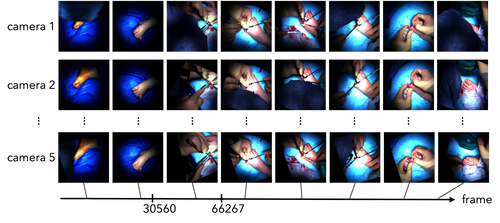}
\vspace{-5mm}
\caption{Video frames after auto-alignment.}
\label{fig:data1}
\vspace{-5mm}
\end{figure}
\section{Experiments and Results}
\label{sec:experiment}
We conducted two experiments to quantitatively and qualitatively investigate our method's efficacy.
\vspace{-3mm}
\\
\noindent
\subsubsection{Dataset.} We used a multi-camera system attached to a surgical light to capture videos of surgical procedures. We captured three types of actual surgical procedures: polysyndactyly, anterior thoracic keloid skin graft, and posttraumatic facial trauma rib cartilage graft. 
From these videos, we prepared five videos which were trimmed to one minute each. Videos 1 and 2 show the surgery of polysyndactyly, videos 3 and 4 show the anterior thoracic keloid skin graft scene, and video 5 shows the surgery of posttraumatic facial trauma rib cartilage graft.
\vspace{-8mm}
\\
\noindent
\subsubsection{Implementation Details.} We used Ubuntu 20.04 LTS OS, an Intel Core i9-12900 for the CPU, and 62GiB of RAM. We defined the area of surgical field as hue ranging from 0 to 30 or from 150 to 179 in HSV color space. 
\vspace{-8mm}
\\
\noindent
\subsubsection{Virtual Single-View Video Generation.} 
\fref{fig:data1} shows a representative frame from the automatic positioning of the surgical video of the polysyndactyly operation using the proposed method. The figure also includes frames with detected camera movement. Once all five viewpoints were aligned, they were fed into the camera-switching algorithm to generate a virtual single-viewpoint video. 
The method requires about 40 minutes every time the cameras move. Please note that it is fully automated and needs no human labor, unlike the existing method, i.e., manual-alignment. 
\vspace{-8mm}
\\
\noindent
\subsubsection{Comparison with Conventional Methods.}
We compared our auto alignment method (auto-alignment) with two conventional methods. In one of these methods, which is used in a hospital camera switching is performed after manual alignment (manual-alignment). The other method switches between camera views with no alignment (no-alignment).

\subsection{Qualitative Evaluation}\label{sec:31} \par
To qualitatively compare our method against baseline methods, we conducted a subjective evaluation. 11 physicians involved in surgical procedures regularly who were expected to actually use the surgical videos were selected as subjects. 

\begin{figure}[tb]
\includegraphics[width=1.0\hsize]{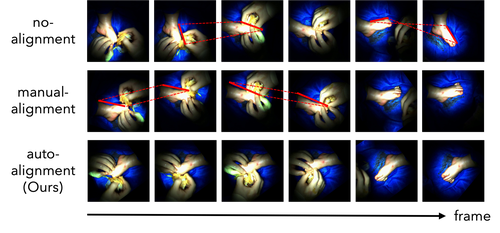}
\vspace{-8mm}
\caption{Video frames obtained using the three methods. The red lines indicate the position and orientation of the instep of patient's foot.}
\label{fig:result}
\vspace{-5mm}
\end{figure}

\fref{fig:result} shows single-view video frames generated by our method and the two conventional methods. 
The red lines indicate the position and orientation of the instep of patient's foot. In video generated with no-alignment, the position and orientation of the insteps changed every time camera switching was performed, making it difficult to observe the surgical region with comfort. Manual-alignment showed better results than no-alignment. However, it was characterized by greater misalignment than the proposed method. It should also be noted that manual alignment requires time and effort. In contrast, our method effectively reduced misalignment between viewpoints even when camera switching was performed.

To perform a qualitative comparison between the three methods, following a previous work~\cite{evalindex1}, we recruited eleven experienced surgeons who were expected to actually use surgical videos and asked them to conduct a subjective evaluation of the captured videos. The subjects were asked to score five statements, 1 (``disagree'') to 5 (``agree'').
\begin{enumerate}[1.]
  \item No discomfort when the cameras switched.
  \item No fatigue, even after long hours of viewing.
  \item Easy to check the operation status.
  \item Easy to see important parts of the frame.
  \item I would like to use this system.
 \end{enumerate}

\begin{figure}[tb]
    \includegraphics[width=1.0\hsize]{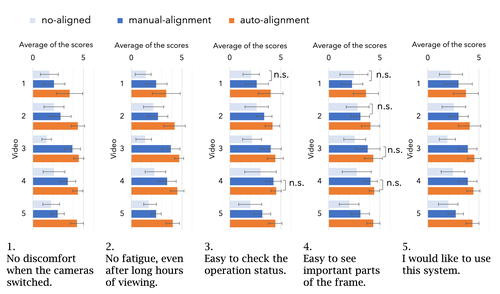}
    \caption{Results of the subjective evaluation experiment.}
    \label{fig:qualitative_evaluation}
    \vspace{-5mm}
\end{figure}

The results are shown in \fref{fig:qualitative_evaluation}.
For almost all videos and statements, the Wilcoxon signed-rank test showed that the proposed method (auto-alignment) scored significantly higher than the two conventional methods. Significant differences are indicated by asterisks in the graphs in \fref{fig:qualitative_evaluation}. The specific p-values are provided in the supplemental material.
The results showing that the proposed method outperformed the conventional methods in statements 1 and 2 suggest that our method generates a stable video with little misalignment between camera viewpoints. Additionally, the results indicating that the proposed method outperformed the conventional method in statements 3 and 4 suggest that the video generated by our method makes it easier to confirm the surgical area. Furthermore, as shown in statement 5, the proposed method received the highest score in terms of the subjects' willingness to use the system in actual medical practice. We believe that the proposed method can contribute to improving the quality of medical care by facilitating stable observation of the surgical field.

Although we observed statistically significant differences between the proposed method and the baselines for almost all the test videos, significant differences were not observed only in Video 3, Statement 3 and Video 4, Statements 3 and 4.
There may be two reasons for this. One is the small number of participants. Since we limited the participants to experienced surgeons, it was quite difficult to obtain a larger sample size. The second reason is differences in the geometry of the surgical fields. Our method is more effective for scenes with a three-dimensional geometry. If the surgical field is flat with fewer feature points, as in the case of the anterior thoracic keloid skin graft procedure, differences between our method and the manual alignment mehod, which does not take into account three-dimensional structures, are less likely to be observed.

\subsection{Quantitative Evaluation}\label{sec:32} \par
Our method aims to reduce the misalignment between viewpoints that occurs when swıtching between multiple cameras and generate single-view surgical videos with less occlusion. To investigate the method's effectiveness, we conducted a quantitative evaluation to assess the degree of misalignment between video frames.

Following a previous work that calculated degree of misalignment between consecutive time-series frames~\cite{evalindex2}, we used two metrics, the interframe transformation fidelity (ITF) and the average speed (AvSpeed).
ITF represents the average peak signal-to-noise ratio (PSNR) between frames as
\begin{equation}
\text{ITF}=\frac{1}{N_{f} - 1} \sum_{i=1}^{N_{f}-1} \text{PSNR}(t)\text{,}
\label{equ:itf}
\end{equation}
where $N_{f}$ is the total number of frames. ITF is higher for videos with less motion blur.
AvSpeed expresses the average speed of feature points. With the total number of frames $N_{f}$ and the number of all feature points in a frame $N_{p}$, AvSpeed is calculated as
\begin{equation}
\text{AvSpeed}=\frac{1}{N_{p}(N_{f} - 1)} \sum_{i=1}^{N_{p}}\sum_{t=1}^{N_{f}-1} \| \dot{z_{i}}(t) \|\text{,}
\label{equ:avspeed}
\end{equation}
where $z_{i}(t)$ denotes the image coordinates of the feature point and is calculated as
\begin{equation}
\dot{z_{i}}(t) = z_{i}(t+1) - z_{i}(t)\text{.}
\label{equ:avspeed2}
\end{equation}

The results are shown in \tref{table:avspeed}.
The ITF of the videos generated using the proposed method was 20\%-50\% higher than that of the videos with manual alignment. The AvSpeed of the videos generated using the proposed method was 40\%–70\% lower than that of the videos with manual alignment, indicating that the shake was substantially corrected.

\begin{table}[tb]
\footnotesize
\centering
  \caption{Results of the quantitative evaluation.}
  \label{table:avspeed}
  \begin{tabular}{ccccccccc}
    \toprule
     & & \multicolumn{3}{c}{ITF [dB] ($\uparrow$)} & & \multicolumn{3}{c}{AvSpeed [pixel/frame] ($\downarrow$)} \\
    \cmidrule{3-5} \cmidrule{7-9}
     Video ID & & \multicolumn{3}{c}{alignment} & & \multicolumn{3}{c}{alignment} \\
       & & no & manual & auto(Ours) & & no & manual & auto(Ours) \\
    \midrule
    1 & & 11.97 & 11.87 & \textbf{17.54} & & 406.3 & 416.1 & \textbf{166.1} \\
    2 & & 11.30 & 11.93 & \textbf{15.77} & & 339.4 & 328.7 & \textbf{195.6} \\
    3 & & 16.17 & 17.85 & \textbf{22.26} & & 448.6 & 230.9 & \textbf{92.2} \\
    4 & & 14.43 & 16.01 & \textbf{19.26} & & 379.0 & 240.7 & \textbf{77.5} \\
    5 & & 15.19 & 17.42 & \textbf{21.66} & & 551.6 & 383.2 & \textbf{169.6} \\
    
    \bottomrule
  \end{tabular}
\end{table}

\section{Conclusion and Discussion}
\label{sec:conclusion}

In this work, we propose a method for generating high-quality virtual single-viewpoint surgical videos captured by multiple cameras attached to a surgical light without occlusion or misalignment through automatic geometric calibration.
In evaluation experiments, we compared our auto-alignment method with manual-alignment and no-alignment. The results verified the superiority of the proposed method both qualitatively and quantitatively. 
The ability to easily confirm the surgical field with the automatically generated virtual single-viewpoint surgical video will contribute to medical treatment.

\paragraph{Limitations.}

Our method relies on visual information to detect the timing of homography calculations (i.e., $t_h$). However, we may use prior knowledge of a geometric constraint such that cameras are at the pentagon corners (\figurename~\ref{fig:surgicallight}).

We assume that the multi-camera surgical light does not move more than twice in 10 minutes for a robust calculation of $D_t$. Although surgeons rarely moved the light more often, fine-tuning the parameter may result in further performance improvement. The current implementation shows misaligned images if the cameras move more frequently.

In the user-involved study, several participants reported noticeable black regions where no camera views were projected. (e.g., \fref{fig:result}). One possible complement is to project pixels from other views.

\subsubsection{Acknowledgements}
This work was partially supported by JSPS KAKENHI Grant Number 22H03617.

\newpage

\bibliographystyle{splncs04}
\bibliography{egbib}

\end{document}